\documentclass[conference]{IEEEtran}
\IEEEoverridecommandlockouts
\usepackage{graphicx,multirow,url,colortbl,hhline,booktabs}
\usepackage{amssymb}
\usepackage{epstopdf}
\usepackage{subfigure}
\usepackage[numbers]{natbib}
\renewcommand{\vec}[1]{\mbox{\boldmath{$#1$}}}
\def\BibTeX{{\rm B\kern-.05em{\sc i\kern-.025em b}\kern-.08em
    T\kern-.1667em\lower.7ex\hbox{E}\kern-.125emX}}
\begin{document}

\title{On dynamic ensemble selection and data preprocessing for multi-class imbalance learning}

\author{\IEEEauthorblockN{Rafael M. O. Cruz}
\IEEEauthorblockA{\textit{Ecole de Technologie Sup\'{e}rieure} \\
Montreal, Canada \\
rafaelmenelau@gmail.com}
\and
\IEEEauthorblockN{Robert Sabourin}
\IEEEauthorblockA{\textit{Ecole de Technologie Sup\'{e}rieure} \\
Montreal, Canada \\
Robert.Sabourin@etsmtl.ca}
\and
\IEEEauthorblockN{George D. C. Cavalcanti}
\IEEEauthorblockA{\textit{Centro de Inform\'{a}tica} \\
\textit{Universidade Federal de Pernambuco}\\
Recife, Brazil \\
gdcc@cin.ufpe.br}
}

\maketitle

\begin{abstract}
	
Class-imbalance refers to classification problems in which many more instances are available for certain classes than for others. Such imbalanced datasets require special attention because traditional classifiers generally favor the majority class which has a large number of instances. Ensemble of classifiers have been reported to yield promising results. However, the majority of ensemble methods applied too imbalanced learning are static ones. Moreover, they only deal with binary imbalanced problems. Hence, this paper presents an empirical analysis of dynamic selection techniques and data preprocessing methods for dealing with multi-class imbalanced problems. We considered five variations of preprocessing  methods and four dynamic selection methods. Our experiments conducted on 26 multi-class imbalanced problems show that the dynamic ensemble improves the F-measure and the G-mean as compared to the static ensemble. Moreover, data preprocessing plays an important role in such cases.

\end{abstract}


\section{Introduction}

Class-imbalance \citep{He09} refers to classification problems in which many more instances are available for certain classes than for others. Particularly, in a two-class scenario, one class contains the majority of instances (the \textit{majority class}), while the other (the \textit{minority class}) contains fewer instances. Imbalanced datasets may originate from real life problems including the detection of fraudulent bank account transactions, telephone calls, biomedical diagnosis, image retrieval and so on.

One of the biggest challenges in imbalanced learning is dealing with multi-class imbalanced problems~\cite{Krawczyk16}. Multi-class imbalanced classification is not as well developed as the binary case, with only a few papers handling this issue~\cite{abdi2016combat,fernandez2013analysing,fernandez2011dynamic}. It is also considered as a more complicated problem, since the relation among the classes is no longer obvious. For instance, one class may be the majority one when compared to some classes, and minority when compared to others. Moreover, we may easily lose performance
on one class while trying to gain it on another~\cite{fernandez2013analysing}.

One way of dealing with imbalanced distributions is to use ensemble learning. As shown in~\cite{Pastor15}, a diverse ensemble can better cope with imbalanced distribution. In particular, Dynamic selection (DS) techniques is seen as an alternative to deal with multi-class imbalance as it explores the local competence of each base classifier according to each new query sample~\cite{CRUZ2018195,Krawczyk16,Roy2018}. Only the base classifiers that attained a certain competence level, in the given local region, are selected to predict the label of the query sample.

A key factor in dynamic selection is the estimation of the classifiers' competences according to each test sample. Usually the estimation of the classifiers competences are based on a set of labeled samples, called the dynamic selection dataset (DSEL). However, As reported in \citep{cruz2016}, dynamic selection performance is very sensitive to the distribution of samples in DSEL. If the distribution of DSEL itself becomes imbalanced, then there is a high probability that the region of competence for a test instance will become lopsided. Thus, the dynamic selection algorithms might end up biased towards selecting base classifiers that are experts for the majority class. With this in mind, we propose the use of data preprocessing methods for training a pool of classifiers as well as balancing the class distribution in DSEL for the DS techniques.

Hence, in this paper, we perform a study on the application of dynamic selection techniques and data preprocessing for handling with multi-class imbalance. Five data preprocessing techniques and four DS techniques as well as static ensemble combination are considered in our experimental analysis. Experiments are conducted using 26 multi-class imbalanced datasets with varying degrees of class imbalance. The following research questions are studied in this paper:

\begin{enumerate}
			\item Does data preprocessing play an important role in the performance of dynamic selection techniques?
			\item Which data preprocessing technique is better suited for dynamic and static ensemble combination?
			\item Do dynamic ensembles present better performance than static ensembles?
\end{enumerate}

This paper is organized as follows: Section~\ref{sec:ds} presents the related works on dynamic selection and describes the DCS and DES methods considered in this analysis. Data preprocessing techniques for imbalance are presented in Section~\ref{sec:preprocessing}. Experiments are conducted in Section~\ref{sec:experiments}. Conclusion and future works are presented in the last section.
 
\section{Dynamic selection}
\label{sec:ds}

A dynamic selection (DS) enables the selection of one or more base classifiers from a pool, given a test instance. This is based on the assumption that each base classifier is an expert in a different local region in the feature space. Therefore, the most competent classifiers should be selected in classifying a new instance. The notion of competence is used in DS as a way of selecting, from a pool of classifiers, the best classifiers to classify a given test instance. Usually, the competence of a base classifier is estimated based on a small region in the feature space surrounding a given test instance, called the \textit{region of competence}. This region is formed using the k-nearest neighbors (KNN) technique, with a set of labeled samples, which can be either the training or validation set. This set is called the Dynamic Selection dataset (DSEL) \citep{CRUZ2018195}. To establish the competence, given a test instance and the DSEL, the literature reports a number of measures classified into individual-based and group-based measures~\cite{CRUZ2018195}. Implementation of several DS techniques can be found on GitHub: ~\url{https://github.com/Menelau/DESlib}~\cite{deslib}.

Among the categories, we focus on the individual-based measures, which consider individual base classifier accuracy for the region of competence. However, the competency measures are calculated differently by different methods in this category. For example, we consider methods in which the competency is measured by pure accuracy \citep{Woods97}, by ranking of classifiers \citep{Sabourin93} or using oracle information \citep{Ko08}. 

Instead of grouping DS strategies by competence measure, we may also group them by selection methodology. Currently, there are two kinds of selection strategies: dynamic classifier selection (DCS) and dynamic ensemble selection (DES). DCS selects a single classifier for a test instance, whereas DES selects an ensemble of classifiers (EoC) to classify a test instance. Both these strategies have been studied in recent years, and some papers are available examining them \citep{Britto14,Cruz15}. In this paper, we evaluate two DCS and two DES strategies: 

\begin{itemize}
	\item \textbf{The Modified Classifier Rank (RANK)} \citep{Sabourin93} is a DCS method that exploits ranks of individual classifiers in the pool for each test instance. The rank of a classifier is based on its local accuracy within a neighborhood of the test instance. More formally, given a test instance assigned to class $C_i$  by a classifier, the ranking of the classifier is estimated as the number of consecutive nearest neighbors assigned to class $C_i$ that have been correctly labeled. The most locally accurate classifier has the highest rank and is selected for classification.
	
	\item \textbf{The Local Class Accuracy (LCA)} \citep{Woods97} estimates the classifier accuracy in a local region around the given test instance and then uses the most locally accurate classifier to classify the test instance. The local accuracy is estimated for each base classifier as the percentage of correct classifications within the local region, but considering only those examples where the classifier predicted the same class as the one it gave for the test instance. The classifier presenting the highest local accuracy is used for the classification of the query sample.
	
	\item  \textbf{The KNORA-Eliminate technique (KNE)}~\cite{Ko08} explores the concept of Oracle, which is the upper limit of a DCS technique. Given the region of competence $\theta_{j}$, only the classifiers that correctly recognize all samples belonging to the region of competence are selected. In other words, all classifiers that achieved a 100\% accuracy in this region (i.e., that are local Oracles) are selected to compose the ensemble of classifiers. Then, the decisions of the selected base classifiers are aggregated using the majority voting rule. If no base classifier is selected, the size of the region of competence is reduced, and the search for the competent classifiers is restarted. 
	
	\item  \textbf{The KNORA-Union technique (KNU)}~\cite{Ko08} selects all classifiers that are able to correctly recognize at least one sample in the region of competence. This method also considers that a base classifier can participate more than once in the voting scheme when it correctly classifies more than one instance in the region of competence. The number of votes of a given base classifier $c_{i}$ is equal to the number of samples in the region of competence, for which it predicted the correct label. For instance, if a given base classifier $c_{i}$ predicts the correct label for three samples belonging to the region of competence, it gains three votes for the majority voting scheme. The votes collected by all base classifiers are aggregated to obtain the ensemble decision.
	
\end{itemize}

These DS techniques are based on different criterion to estimate the local competence of the base classifiers. For example, while both the RANK and the LCA are DCS strategies, the former measures the competence based on ranking, and the latter based on classifier accuracy. On the other hand, the two DES strategies (KNE and KNU) are based on Oracle information. These techniques were selected as they were among the best performing DS methods according to the experimental analysis conducted in~\cite{CRUZ2018195}.

Nevertheless, a crucial aspect in the performance of the dynamic selection techniques is the distribution of the dynamic selection dataset (DSEL), as the local competence of the base classifiers are estimated based on this set. Hence, preprocessing techniques can really benefit DS techniques as they can be employed to edit the distribution of DSEL, prior to performing dynamic selection. 

\section{Data preprocessing}
\label{sec:preprocessing}
		
		Changing the distribution of the training data to compensate for poor representativeness of the minority class is an effective solution for imbalanced problems, and a plethora of methods are available in this regards. Branco et al.~\cite{Branco16} divided such methods into three categories, namely, stratified sampling, synthesizing new data, and combinations of the two previous methods. While the complete taxonomy is available in \citep{Branco16}, we will center our attention on the methods that have been used together with ensemble learning~\cite{Pastor15}. 
		
		One important category is under-sampling, which removes instances from the majority class to balance the distribution.
		Random under-sampling (RUS) \citep{Barandela03} is one such method. RUS has been coupled with boosting (RUSBoost) \citep{Seiffert10} and with Bagging \citep{Barandela03}. A major drawback of RUS is that it can discard potentially useful data which can be a problem when using dynamic selection approaches.
		
		The other strategy is the generation of new synthetic data. Synthesizing new instances has several known advantages \citep{Chawla02}, and a wide number of proposals are available for building new synthetic examples. In this context, a famous method that uses interpolation to generate new instances is SMOTE~\cite{Chawla02}. SMOTE over-samples the minority class by generating new synthetic data. A number of methods have been developed based on the principle of SMOTE, such as, Borderline-SMOTE \citep{Han05}, ADASYN \citep{He08}, RAMO \citep{Chen10} and Random balance \citep{Pastor15RB}. Furthermore, Garcia et al.~\cite{Garcia12} observed that over-sampling consistently outperforms under-sampling for strongly imbalanced datasets. 
		
		Hence, in this work we considered three over-sampling techniques. Similar to~\cite{abdi2016combat}, the class with the highest number of examples is considered the majority class, while all others are considered minority classes. Then, the over-sampling techniques are applied to generate synthetic samples for each minority class.
		
		\begin{itemize}
			\item \textbf{Synthetic Minority Over-sampling Technique (SMOTE)} \cite{Chawla02}, which creates artificial instances for the minority class. The process works as follows: Let $\vec{x}_i$ be an instance from the minority class. To create an artificial instance from $\vec{x}_i$, SMOTE first isolates the k-nearest neighbors of $\vec{x}_i$, from the minority class. Afterward, it randomly selects one neighbor and randomly generates a synthetic example along the imaginary line connecting $\vec{x}_i$ and the selected neighbor.
			
			\item \textbf{Ranked Minority Over-sampling (RAMO)} \citep{Chen10}, which performs a sampling of the minority class according to a probability distribution, followed by the creation of synthetic instances. The RAMO process works as follows: For each instance $\vec{x}_i$ in the minority class, its $k_1$ nearest neighbors ($k_1$ is a user defined neighborhood size) from the whole dataset are isolated. The weight $r_i$ of $\vec{x}_i$ is defined as:
			\begin{equation}\label{eqn:RAMO_Weight}
			r_i = \frac{1}{1+exp(-\alpha.\delta_i)},
			\end{equation}
			where $\delta_i$ is the number of majority cases in the k-nearest neighborhood. Evidently, an instance with a large weight indicates that it is surrounded by majority class samples, and thus difficult to classify.
			
			After determining all weights, the minority class is sampled using these weights to get a sampling minority dataset $G$. The synthetic samples are generated for each instance in $G$ by using SMOTE on $k_2$ nearest neighbors where $k_2$ is a user-defined neighborhood size.
			
			\item \textbf{Random Balance (RB)} \citep{Pastor15RB}, which relies on the amount of under-sampling and over-sampling that is problem specific and that has a significant influence on the performance of the classifier concerned. RB maintains the size of the dataset, but varies the proportion of the majority and minority classes, using a random ratio. This includes the case where the minority class is over represented and the imbalance ratio is inverted. Thus, repeated applications of RB produce datasets having a large imbalance ratio variability, which promotes diversity~\cite{Pastor15RB}. SMOTE and random under-sampling are used to respectively increase or reduce the size of the classes to achieve the desired ratios. 
			
			Given a dataset $S$, with minority class $S_P$ and majority class $S_N$, the RB procedure can be described as follows:
			
			\begin{enumerate}
				\item The modified size, $newMajSize$, of the majority class, is defined by a random number generated between 2 and $|S| - 2$ (both inclusive). Accordingly, the modified size, $newMinSize$, of the minority class becomes $|S| - newMajSize$.
				
				\item If $newMajSize < |S_N|$, the majority class $S'_N$ is created by RUS the original $S_N$ so that the final size $|S'_N| = newMajSize$. Consequently, the new minority class $S'_P$ is obtained from $S_P$ using SMOTE to create $newMinSize-|S_P|$ artificial instances.
				
				\item Otherwise, $S'_P$ is the class created by RUS $S_P$. On the other hand, $S'_N$ is the class that includes artificial samples generated using SMOTE on $S_N$. Thus, finally, $|S'_P| = newMinSize$ and $|S'_N| = newMajSize$.
			\end{enumerate}
		\end{itemize}

\section{Experiments}
\label{sec:experiments}
\subsection{Datasets}

				A total of 26 multi-class imbalanced datasets taken from the Keel repository~\cite{Fdez11} was used in this analysis. The key features of the datasets are presented in Table~\ref{table:multiclassdatasets}. The IR is computed as the proportion of the number of the majority class examples to the number of minority class examples. In this case, the class with maximum number of examples is the majority class, and the class with the minimum number of examples is the minority one. We grouped the datasets according to their IRs using the group definitions suggested by~\cite{Alberto08}. Datasets with low IR ($IR < 3$) are highlighted with dark gray, whereas datasets with medium IR ($3 < IR < 9$) are in light gray.
				
				\begin{table*}[htbp]
					\centering
					\caption{Characteristics of the 26 multi-class imbalanced datasets taken from the Keel repository. Column \#E shows the number of instances in the dataset, column \#A the number of attributes (numeric/nominal), \#C shows the number of classes in the dataset, and column IR the imbalance ratio.}
					\label{table:multiclassdatasets} 
					\resizebox{0.750\textwidth}{!}{
						\begin{tabular}{c c c c c c c c c c}
							\hline
							\textbf{Dataset} & \textbf{\#E} & \textbf{\#A} & \textbf{\#C} & \textbf{IR} & \textbf{Dataset} & \textbf{\#E} & \textbf{\#A} & \textbf{\#C} & \textbf{IR}\\
							\midrule
							\rowcolor[gray]{0.3}					
							\textcolor{white}{Vehicle} & \textcolor{white}{846} & \textcolor{white}{(18/0)} & \textcolor{white}{4} & \textcolor{white}{1.09} & \cellcolor{white} CTG & \cellcolor{white} 2126 & \cellcolor{white}  (21/0) & \cellcolor{white} 3 & \cellcolor{white}  9.40 \\
							\rowcolor[gray]{0.3}					
							\textcolor{white}{Wine} & \textcolor{white}{178} & \textcolor{white}{(13/0)} & \textcolor{white}{3} & \textcolor{white}{1.48} & \cellcolor{white} Zoo & \cellcolor{white}101 &\cellcolor{white} (16/0) &\cellcolor{white} 7 &\cellcolor{white} 10.25\\
							\rowcolor[gray]{0.3}
							\textcolor{white}{Led7digit} & \textcolor{white}{500} & \textcolor{white}{(7/0)} & \textcolor{white}{10} & \textcolor{white}{1.54} &\cellcolor{white} Cleveland & \cellcolor{white} 467 & \cellcolor{white} (13/0) & \cellcolor{white} 5 & \cellcolor{white} 12.62\\
							\rowcolor[gray]{0.3}					
							\textcolor{white}{Contraceptive} & \textcolor{white}{1473} & \textcolor{white}{(9/0)} & \textcolor{white}{3} & \textcolor{white}{1.89} & \cellcolor{white} Faults & \cellcolor{white} 1941 & \cellcolor{white} (27/0) & \cellcolor{white} 7 & \cellcolor{white} 14.05\\
							\rowcolor[gray]{0.3}
							\textcolor{white}{Hayes-Roth} & \textcolor{white}{160} & \textcolor{white}{(4/0)} & \textcolor{white}{3} & \textcolor{white}{2.10} & \cellcolor{white} Autos & \cellcolor{white} 159 & \cellcolor{white} (16/10) & \cellcolor{white} 6 & \cellcolor{white} 16.00\\
							\rowcolor[gray]{0.3}
							\textcolor{white}{Column3C} & \textcolor{white}{310} & \textcolor{white}{(6/0)} & \textcolor{white}{3} & \textcolor{white}{2.50} & \cellcolor{white} Thyroid & \cellcolor{white} 7200 & \cellcolor{white} (21/0) & \cellcolor{white} 3 & \cellcolor{white} 40.16\\
							\rowcolor[gray]{0.3}
							\textcolor{white}{Satimage} & \textcolor{white}{6435} & \textcolor{white}{(36/0)} & \textcolor{white}{7} &\textcolor{white}{2.45} & \cellcolor{white} Lymphography & \cellcolor{white} 148 & \cellcolor{white} (3/15) & \cellcolor{white} 4 & \cellcolor{white} 40.50\\
							\rowcolor[gray]{0.7}
							Laryngeal3 & 353 & (16/0) & 3 & 4.19 & \cellcolor{white} Post-Operative & \cellcolor{white} 87 & \cellcolor{white} (1/7) & \cellcolor{white} 3 & \cellcolor{white} 62.00\\									\rowcolor[gray]{0.7}
							New-thyroid & 215 & (5/0) & 3 & 5.00 & \cellcolor{white} Wine-quality red & \cellcolor{white} 1599 & \cellcolor{white}(11/0) & \cellcolor{white} 11 &  \cellcolor{white}68.10\\
							\rowcolor[gray]{0.7}
							Dermatology & 358 & (33/0) & 6 & 5.55  & \cellcolor{white} Ecoli & \cellcolor{white} 336 & \cellcolor{white} (7/0) &\cellcolor{white} 8 &\cellcolor{white} 71.50\\
							\rowcolor[gray]{0.7}
							Balance & 625 & (4/0) & 3 & 5.88 & \cellcolor{white} Page-blocks &\cellcolor{white} 5472 & \cellcolor{white} (10/0) & \cellcolor{white} 5 & \cellcolor{white} 175.46\\
							\rowcolor[gray]{0.7}
							Flare & 1066 & (0/11) & 6 & 7.70 & \cellcolor{white} Abalone & \cellcolor{white} 4139 & \cellcolor{white} (7/1) & \cellcolor{white} 18 & \cellcolor{white} 45.93\\
							\rowcolor[gray]{0.7}
							Glass & 214 & (9/0) & 6 & 8.44 & \cellcolor{white} Nursery & \cellcolor{white} 12690 & \cellcolor{white} (0/8) & \cellcolor{white} 5 &  \cellcolor{white}2160.00\\
							
							\hline
						\end{tabular}}
					\end{table*} 

\subsection{Experimental setup}

The Weka 3.8 along with Matlab 8.4.0 was used in the experiments. Results were obtained with a $5\times 2$ stratified cross-validation. Performance evaluation is conducted using the multi-class generalization of the AUC, F-measure and G-mean, as the standard classification accuracy is not suitable for imbalanced learning~\cite{Pastor15}. 

The pool size for all ensemble techniques was set to 100. The classifier used as a base classifier in all ensembles was J48, which is the Java implementation of Quinlan's C4.5, available in Weka 3.8. Here, C4.5 was used with Laplace smoothing at the leaves, but without pruning and collapsing as recommended in~\cite{Pastor15}.

All preprocessing techniques were combined with Bagging during the pool generation phase. Table \ref{tbl:EnsembleMethods} lists such combinations. The preprocessing techniques, RAMO and SMOTE, have user-specified parameters. In the case of RAMO, we used $k_1 = 10$, $k_2 = 5$ and $\alpha = 0.3$. For SMOTE and RB, the number of nearest neighbors was 5. These parameter settings were adopted from \citep{Pastor15}. Finally, for all the dynamic selection methods, we used 7 nearest neighbors to define the region of competence as in \citep{Cruz15,CRUZ2018195}.

\begin{table}[htp]
	\centering
	\footnotesize
	\caption{Preprocessing methods used for classifier pool generation.}\label{tbl:EnsembleMethods}
	\resizebox{0.45\textwidth}{!}{  
	\begin{tabular}{lll}
		\hline
		\multicolumn{3}{l}{Bagging based methods} \\
		\hline
		Abbr. & Name & Description \\
		\hline
		Ba & Bagging & Bagging without preprocessing\\
		Ba-RM100 & Bagging+RAMO 100\%& RAMO to double the minority class \\
		Ba-RM & Bagging+RAMO & RAMO to make equal size for both classes \\
		Ba-SM100 & Bagging+SMOTE 100\% & SMOTE to double the minority class \\
		Ba-SM & Bagging+SMOTE & SMOTE to make equal size for both classes \\
		Ba-RB & Bagging+RB & RB to randomly balance the two classes \\
		\hline
	\end{tabular}}
\end{table}

\begin{figure}[htp]

			\centering
			\includegraphics[width=0.450\textwidth]{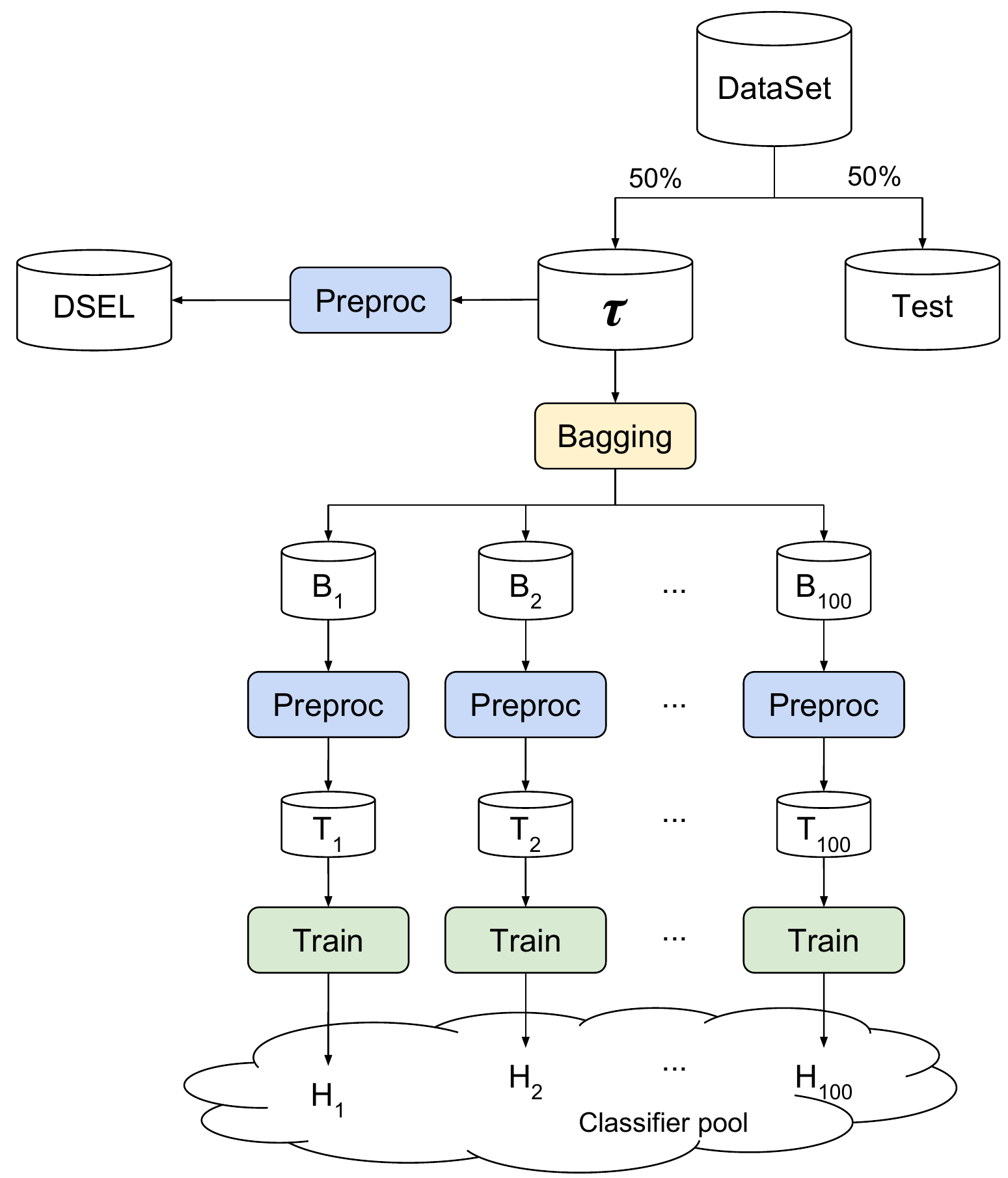}\\
			\caption{The framework for training base classifiers and to prepare a DSEL for testing. Here, $\vec{\tau}$ is the training data derived from the original dataset, $B_i$ is the dataset generated from the $i$th Bagging iteration, $T_i$ is the dataset produced by preprocessing (\textit{Preproc}) $B_i$ and $H_i$ is the $i$th base classifier.}\label{fig:OneFold}
		\end{figure}

The complete framework for a single replication is presented in Figure \ref{fig:OneFold}. The original dataset was divided into two equal halves. One of them was set aside for testing, while the other half was used to train the base classifiers and to derive the dynamic selection set. Let us now highlight the process of setting up the DSEL. Here, instead of dividing the training set, we augment it using the data preprocessing, to create DSEL. Moreover, the Bagging method is applied to the training set, generating a bootstrap with 50\% of the data. Then, the preprocessing method is applied to each bootstrap, and the resulting dataset is used to generate the pool of classifiers. Since we considered a single training dataset, the DSEL dataset has an overlap with the datasets used during Bagging iterations. However, the randomized nature of the preprocessing methods allows the DSEL not to be exactly the same as the training datasets. Thus, avoiding overfitting issues.

\subsection{Results according to data preprocessing method}

In this section, we compare the performance of each preprocessing method with respect to each ensemble technique. Tables~\ref{table:rankAUC}, \ref{table:rankF} and \ref{table:rankG} show the average rank for the AUC, F-measure and G-mean, respectively. The best average rank is in bold. We can see that the SM and SM100 obtained the best results. Furthermore, the configuration using only Bagging always presented the highest average rank. 

The Finner's \citep{Finner93} step-down procedure was conducted at a $95\%$ significance level to identify all methods that were equivalent to the best ranked one. The analysis demonstrates that considering the F-measure and G-mean, the result obtained using preprocessing techniques is always statistically better when compared to using only Bagging.

\begin{table}[h!] 
\centering 
\caption{Average rankings according to AUC. Methods in brackets are statistically equivalent to the best one.} 
\label{table:rankAUC}  
\resizebox{0.45\textwidth}{!}{  
\begin{tabular}{c c c c c c c}  
\hline  
 
Algorithm & Bagging & RM & RM100 & SM & SM100 & RB \\ \hline  
 
KNE & 3.85 & 3.81 & [2.88] & [3.46] & \textbf{2.62} & 4.38 \\ 
KNU & [3.35] & 3.50 & \textbf{2.88} & 3.73 & [3.27] & 4.27 \\ 
LCA & [3.23] & [3.19] & [3.42] & 3.77 & \textbf{2.77} & 4.62 \\ 
RANK & 3.73 & [3.38] & 3.92 & [3.42] & \textbf{2.88} & [3.65] \\ 
STATIC & [3.42] & 3.50 & \textbf{2.58} & 3.81 & [3.31] & 4.38 \\ 
\hline  
 
\end{tabular}} 
\end{table}
\begin{table}[h!] 
\centering 
\caption{Average rankings according to F-measure. Methods in brackets are statistically equivalent to the best one.} 
\label{table:rankF}  
\resizebox{0.45\textwidth}{!}{  
\begin{tabular}{c c c c c c c}  
\hline  
 
Algorithm & Bagging & RM & RM100 & SM & SM100 & RB \\ \hline  
 
KNE & 5.00 & [3.35] & [3.15] & [3.23] & \textbf{2.65} & 3.62 \\ 
KNU & 4.42 & \textbf{3.08} & [3.42] & [3.35] & [3.19] & [3.54] \\ 
LCA & 3.92 & 3.42 & 3.62 & [3.00] & \textbf{2.38} & 4.65 \\ 
RANK & 4.19 & [3.46] & 3.69 & [3.38] & \textbf{2.54} & 3.73 \\ 
STATIC & 4.00 & [3.54] & [3.15] & [3.46] & \textbf{2.81} & 4.04 \\ 
\hline  
 
\end{tabular}} 
\end{table}
\begin{table}[h!] 
\centering 
\caption{Average rankings according to G-mean. Methods in brackets are statistically equivalent to the best one.} 
\label{table:rankG}  
\resizebox{0.45\textwidth}{!}{  
\begin{tabular}{c c c c c c c}  
\hline  
 
Algorithm & Bagging & RM & RM100 & SM & SM100 & RB \\ \hline  
 
KNE & 5.42 & [3.46] & [3.50] & \textbf{2.81} & [3.00] & [2.81] \\ 
KNU & 5.00 & [3.00] & 4.00 & \textbf{2.81} & [4.00] & 2.19 \\ 
LCA & 4.38 & [3.46] & 4.08 & \textbf{2.69} & [3.00] & [3.38] \\ 
RANK & 4.73 & [3.42] & 3.92 & [3.27] & \textbf{2.54} & [3.12] \\ 
STATIC & 3.92 & [3.50] & [3.50] & \textbf{3.15} & [3.54] & [3.38] \\ 
\hline  

\end{tabular}} 
\end{table}

\begin{figure*}[htbp]
	\centering
	\subfigure[AUC]{\includegraphics[width=2.2in]{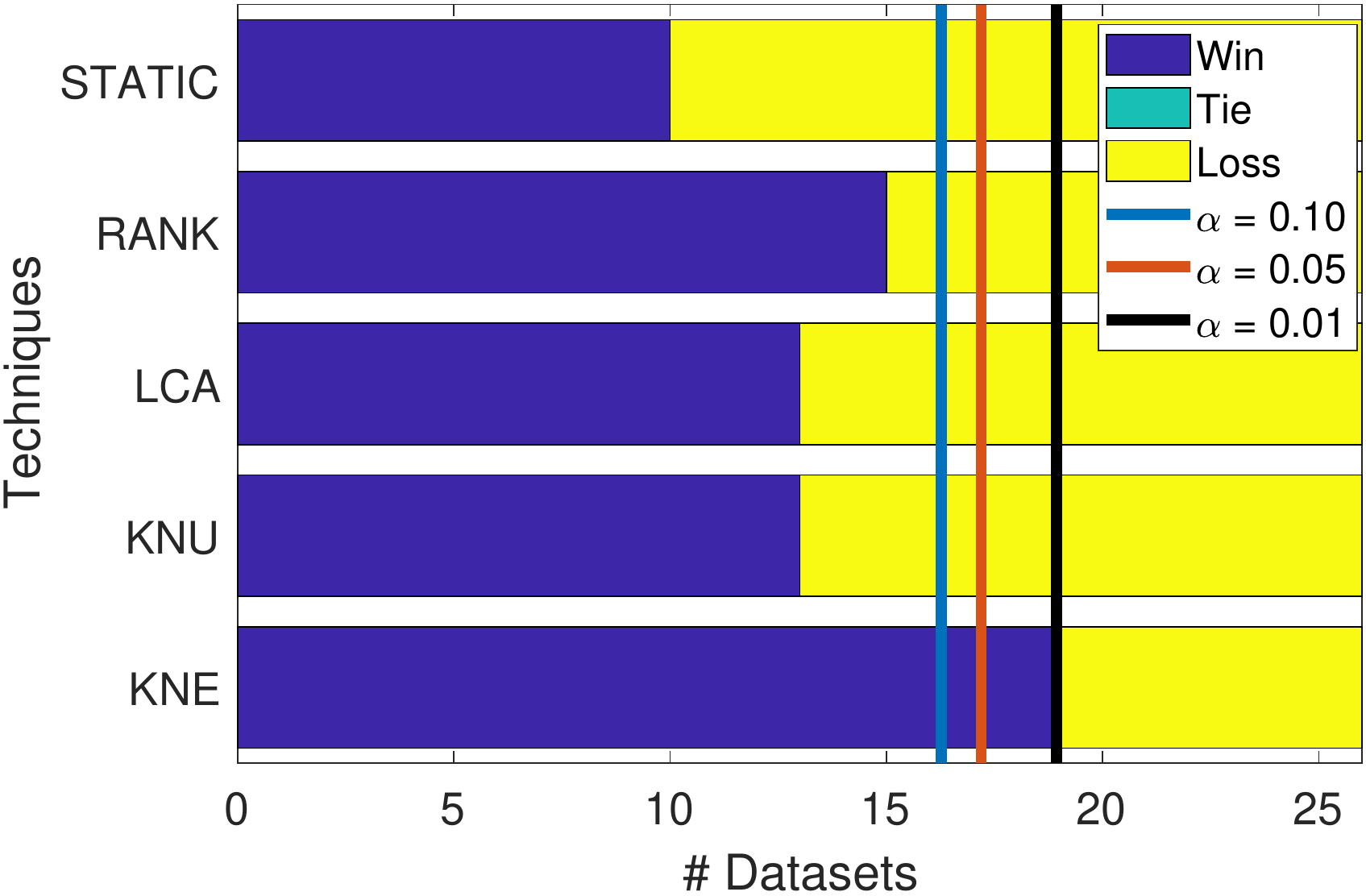}}   
	\subfigure[F-measure]{\includegraphics[width=2.2in]{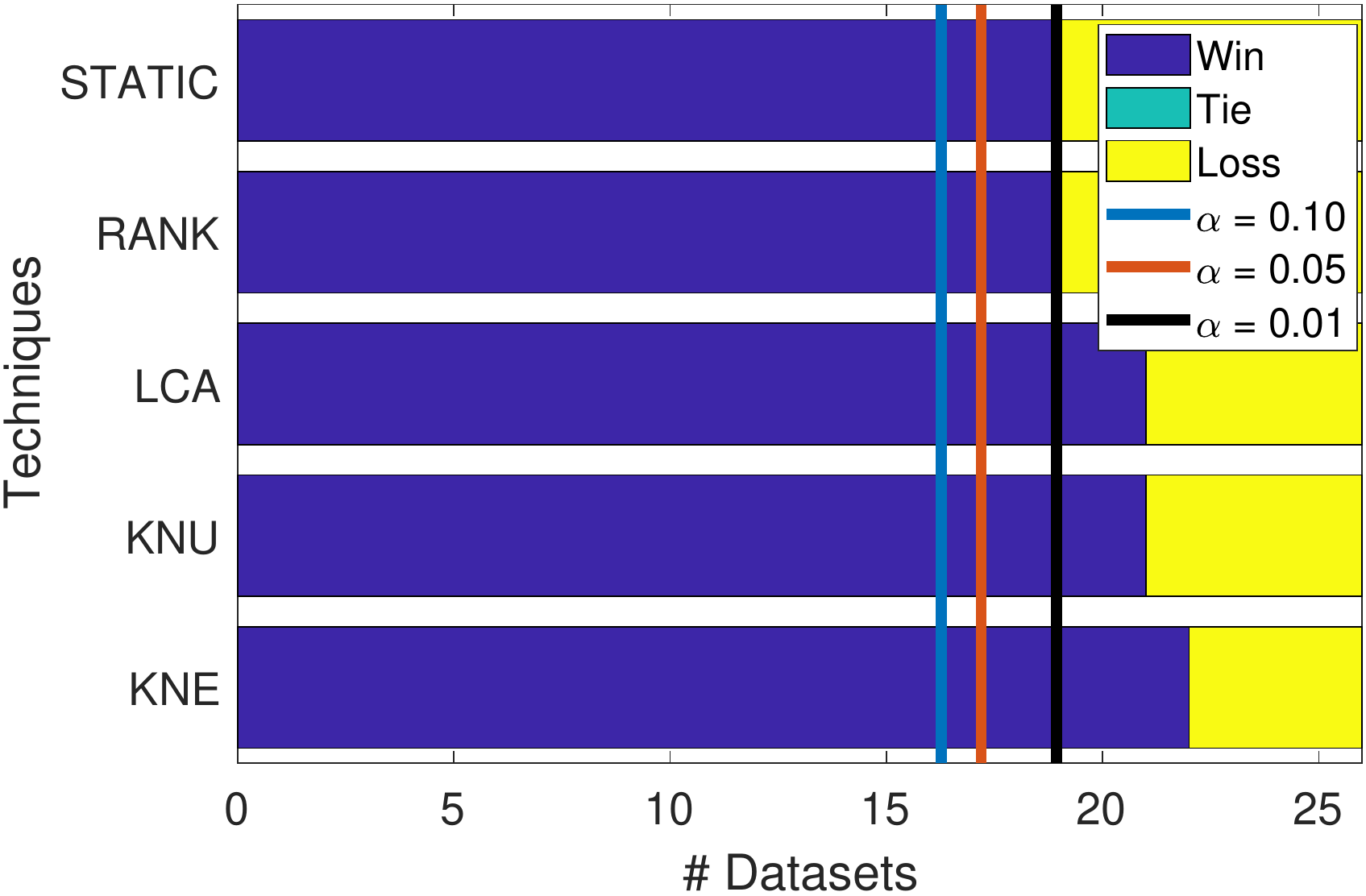}}	
	\subfigure[G-mean]{\includegraphics[width=2.2in]{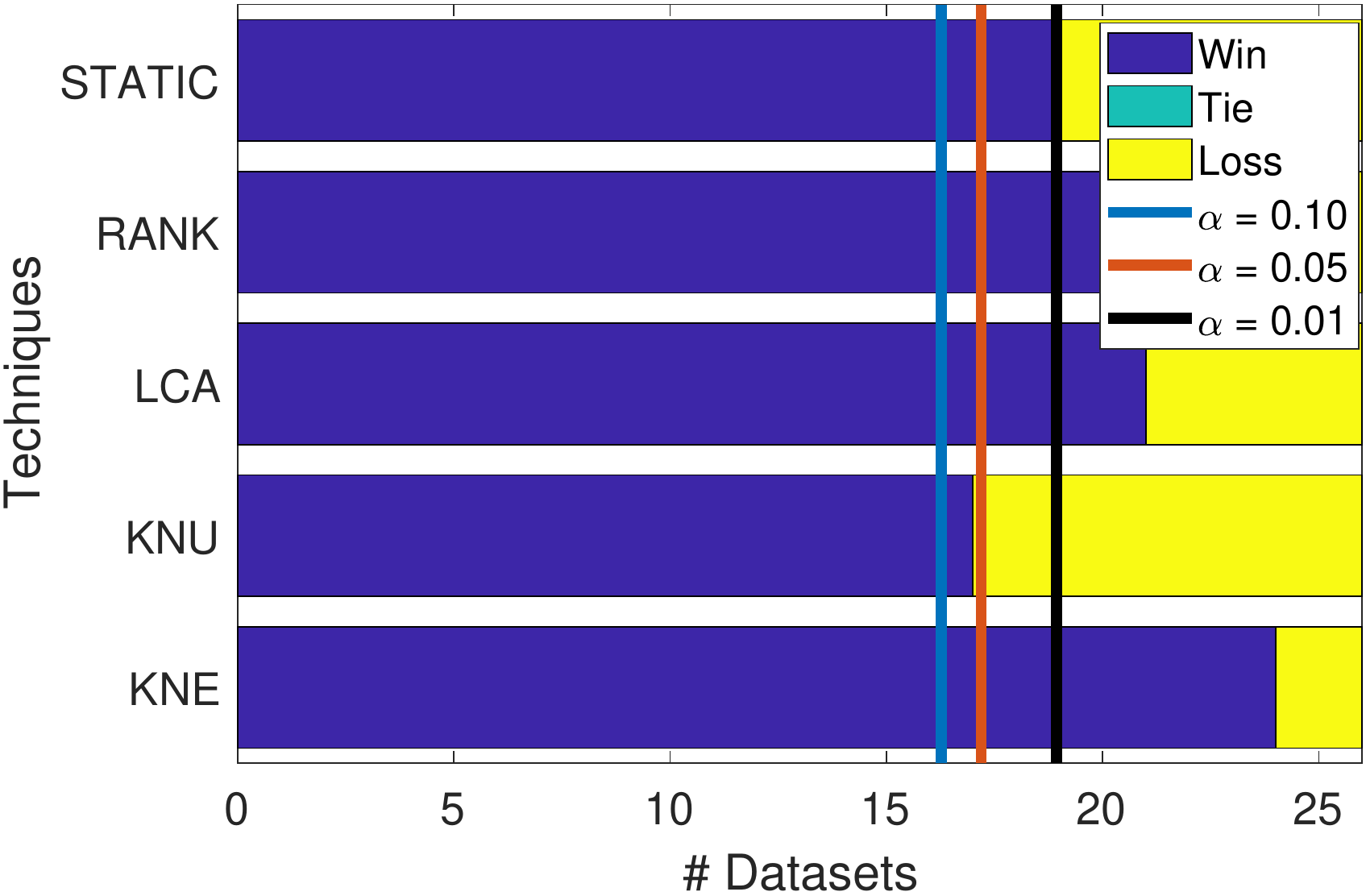}}   
	
	\caption{Sign test computed over the wins, ties and losses. The vertical lines represents the critical value for at a significance level $\alpha = \left\lbrace 0.1, 0.05, 0.01 \right\rbrace $.}
	\label{fig:wintielosDES}
	
\end{figure*}	

Moreover, we conducted a pairwise comparison between each ensemble methods using data preprocessing with the same methods using only Bagging (baseline). For the sake of simplicity, only the best data preprocessing for each technique was considered (i.e., the best result of each row of Tables  \ref{table:rankAUC}, \ref{table:rankF} and \ref{table:rankG}). The pairwise analysis is conducted using the Sign test, calculated on the number of wins, ties, and losses obtained by each method using preprocessing techniques, compared to the baseline. The results of the Sign test is presented in Figure~\ref{fig:wintielosDES}.

The Sign test demonstrated that the data preprocessing significantly improved the results of these techniques according to the F-measure and G-mean. Considering these two metrics, all techniques obtained a significant number of wins for a significance level $\alpha = 0.05$. Moreover, three out of five techniques presented a significant number of wins for $\alpha = 0.01$. Hence, the results obtained demonstrate that data preprocessing techniques indeed play an important role when dealing with multi-class imbalanced problems. 

Furthermore, DS techniques are more benefited from the application of data preprocessing (i.e., presented a higher number of wins). This results can be explained by the fact the data preprocessing techniques are applied in two stages: First, it is used in the ensemble generation stage in order to generate a diverse pool of classifiers. Then, they are also used in order to balance the distribution of the dynamic selection dataset for the estimation of the classifiers' competences.

\subsection{Dynamic selection vs static combination}

In this experiment we compare the performance of the dynamic selection approaches versus static ones. For each technique, the best performing data preprocessing technique is selected (i.e., best result from each row of Tables \ref{table:rankAUC}, \ref{table:rankF} and \ref{table:rankG}). Then, new average ranks are calculated for these methods. Table~\ref{table:ranks} presents the average rank of the top techniques according to each metric.
					
						\begin{table}[htbp]
							\centering
							\footnotesize
							\centering
							\label{table:ranks}
							\caption{Average ranks for the best ensemble methods. (a) According to AUC, (b) according to F-measure and (c) according to G-mean. Results that are statistically equivalent to the best one are in brackets.}\label{tbl:Compare_allmulti}
							\resizebox{.45\textwidth}{!}{
								\begin{tabular}{@{\extracolsep{9pt}}@{\kern\tabcolsep}l@{}r@{}l@{}r@{}l@{}r@{\kern\tabcolsep}}
									\toprule
									\multicolumn{2}{l}{(a) AUC} & \multicolumn{2}{l}{(b) F-measure} & \multicolumn{2}{l}{(c) G-mean} \\
									\cmidrule{1-2}\cmidrule{3-4}\cmidrule{5-6}
									Methods & Rank & Method & Rank & Method & Rank \\
									
									\midrule
									
									Ba-SM+KNU &  2.04 & Ba-RM+KNU & 2.15 &Ba-SM+KNE &  2.23  \\ 
									Ba-SM100+KNE &  [2.42] & Ba-SM100+KNE & [2.31] &Ba-SM+KNU &  [2.42]  \\ 
									Ba-SM &  2.50 & Ba-SM100 & [2.46] & Ba-SM &  [2.62]  \\ 
									Ba-SM100+RANK &  3.77 & Ba-SM100+RANK & 3.58 &Ba-SM100+RANK &  3.38  \\ 
									Ba-SM100+LCA &  4.27 & Ba-SM100+LCA & 4.50 &Ba-SM+LCA &  4.35  \\ 
									
									\bottomrule
								\end{tabular}}
							\end{table}

Based on the average ranks, we can see that the DES techniques present a lower average rank when compared to that of the static combination for the three performance measures. Hence, DES techniques are suitable for dealing with multi-class imbalance. The performance of DES techniques (KNE and KNU) and the static combination were statistically equivalent considering the F-measure and G-mean, while the performance of the KNU was significantly better considering the AUC. On the other hand, the DCS techniques (LCA and RANK) presented a higher average rank when compared to the static ensemble, and may not be suitable to handle multi-class imbalanced problems.

\section{Conclusion}

In this work, we conducted a study on dynamic ensemble selection and data preprocessing for solving the multi-class imbalanced problems. A total of four dynamic selection techniques and five preprocessing techniques were evaluated in this experimental study.

Results obtained over 26 multi-class imbalanced problems demonstrate that the dynamic ensemble selection techniques studied (KNE and KNU) obtained a better result than static ensembles based on AUC, F-measure and G-mean. Moreover, the use of data preprocessing significantly improves the performance of DS and static ensembles. In particular, the SMOTE technique presented the best results. Furthermore, DS techniques seems to benefit more of data preprocessing methods since they are applied not only to generate the pool of classifiers but also to edit the distribution of the dynamic selection dataset.

Future works would involve the definition of new pre-processing techniques specific to deal with multi-class imbalance as well as the definition of cost-sensitive dynamic selection techniques to handle multi-class imbalanced problems.

\bibliographystyle{IEEEtran}
\bibliography{ThesisBib}

\end{document}